\documentclass[letterpaper]{article} 
\usepackage{aaai2026}  
\usepackage{times}  
\usepackage{helvet}  
\usepackage{courier}  
\usepackage[hyphens]{url}  
\usepackage{graphicx} 
\usepackage{natbib}  
\usepackage{caption} 
\usepackage{algorithm}
\usepackage{algorithmic}

\usepackage{amsmath}
\usepackage{amssymb}  

\usepackage{graphicx}
\usepackage{subcaption}

\usepackage{makecell,amsmath}
\usepackage{newfloat}
\usepackage{listings}
\DeclareCaptionStyle{ruled}{labelfont=normalfont,labelsep=colon,strut=off} 
\floatstyle{ruled}
\newfloat{listing}{tb}{lst}{}
\floatname{listing}{Listing}
 \nocopyright 

\title{Beyond Error-Based Optimization: Experience-Driven Symbolic Regression with Goal-Conditioned Reinforcement Learning}

\author{
  Jianwen Sun\textsuperscript{\rm 1,2},
  Xinrui Li\textsuperscript{\rm 1,2},
  Fuqing Li\textsuperscript{\rm 3},
  Xiaoxuan Shen\textsuperscript{\rm 1,2}\thanks{Corresponding author.}
}

\affiliations{
  \textsuperscript{\rm 1}Faculty of Artificial Intelligence in Education, Central China Normal University, Wuhan 430079, China\\
  \textsuperscript{\rm 2}National Engineering Research Center for Educational Big Data, Central China Normal University, Wuhan 430079, China\\
  \textsuperscript{\rm 3}Institute of Collaborative Innovation, University of Macau, Macau 999078, China
}

\usepackage{bibentry}

\begin{document}

\maketitle  

\begin{abstract}

Symbolic Regression aims to automatically identify compact and interpretable mathematical expressions that model the functional relationship between input and output variables. Most existing search-based symbolic regression methods typically rely on the fitting error to inform the search process. However, in the vast expression space, numerous candidate expressions may exhibit similar error values while differing substantially in structure, leading to ambiguous search directions and hindering convergence to the underlying true function. To address this challenge, we propose a novel framework named EGRL-SR (Experience-driven Goal-conditioned Reinforcement Learning for Symbolic Regression). In contrast to traditional error-driven approaches, EGRL-SR introduces a new perspective: leveraging precise historical trajectories and optimizing the action-value network to proactively guide the search process, thereby achieving a more robust expression search. Specifically, we formulate symbolic regression as a goal-conditioned reinforcement learning problem and incorporate hindsight experience replay, allowing the action-value network to generalize common mapping patterns from diverse input-output pairs. Moreover, we design an all-point satisfaction binary reward function that encourages the action-value network to focus on structural patterns rather than low-error expressions, and concurrently propose a structure-guided heuristic exploration strategy to enhance search diversity and space coverage. Experiments on public benchmarks show that EGRL-SR consistently outperforms state-of-the-art methods in recovery rate and robustness, and can recover more complex expressions under the same search budget. Ablation results validate that the action-value network effectively guides the search, with both the reward function and the exploration strategy playing critical roles.

\end{abstract}


\section{Introduction}

Symbolic Regression (SR) aims to discover compact and interpretable mathematical expressions that capture the underlying functional relationships between input and output variables. Given a noise-free dataset $D = \{(x_i, y_i)\}_{i=1}^N$, where $x_i$ denotes the input feature and $y_i$ the target value, the objective of SR is to identify a function $f$ in a concise closed-form analytical expression such that $y_i = f(x_i)$ holds for every $i \in \{1, \dots, N\}$. Unlike traditional regression methods that rely on pre-specified functional forms, SR conducts flexible search in a vast space of symbolic expressions without requiring prior assumptions, making it widely applicable in scientific discovery and engineering modeling. However, as the length of expressions increases, the search space expands exponentially, and this problem has been proven to be NP-hard \cite{1}.

In recent years, advances in neural networks for sequence modeling have encouraged researchers to develop end-to-end methods for expression generation\cite{2,3,4,5}. While these approaches improve fitting accuracy through large-scale pretraining, the expression space is virtually unbounded, making it difficult to learn a stable and generalizable distribution. Consequently, the generated expressions often exhibit structurally implausible forms and contain redundant symbols.

Therefore, in the task of accurately recovering complex expressions, researchers increasingly favor search-based symbolic regression methods that incorporate adaptive mechanisms. Representative approaches include: Genetic Programming, which employs the fitting error as a fitness function to evolve expression structures via evolutionary operators \cite{6,7,8}; the Equation Learner method, which encodes expressions using special neural networks and jointly optimizes both their structure and parameters by minimizing the fitting error  \cite{9,10}; Monte Carlo Tree Search, each node represents a symbolic expression in prefix form, and the algorithm prioritizes exploring branches that are expected to reduce the fitting error \cite{11,12}; and Reinforcement Learning approaches, which formulate expression generation as a sequential decision-making process where a policy network constructs expressions guided by reward signals derived from fitting accuracy \cite{13,14}.

In summary, most existing search-based symbolic regression methods rely heavily on fitting error to guide the search direction, yet this strategy exhibits clear limitations in complex expression spaces. As illustrated in  Figure \ref{figure2}, within the interval $[-1, 1]$, many structurally distinct expressions can achieve low fitting error, while exhibiting drastically different behaviors outside the interval. Optimizing solely for error often leads to divergent search trajectories without clear direction, and once the search veers off course, the algorithm struggles to backtrack or recover. Consequently, the final expressions tend to fall into one of two undesirable extremes: either they are concise but insufficient accuracy, or excessively complex in structure to compensate for error—both of which violate the fundamental objectives of symbolic regression.

To overcome the limitations of existing symbolic regression methods that overly rely on fitting error for search guidance, we propose a novel framework—Experience-driven Goal-conditioned Reinforcement Learning for Symbolic Regression (EGRL-SR). Built upon Goal-Conditioned Reinforcement Learning (GCRL) \cite{28} and enhanced with Hindsight Experience Replay (HER)\cite{29}, EGRL-SR differs fundamentally from error-driven approaches by learning from historical construction trajectories. It generalizes $x-y$ transformation patterns across different targets $y$ to improve expression generation and actively guide the search process. By focusing on reusable structural patterns instead of local optima tied to specific targets, the method effectively mitigates the pitfalls of fitting-error-based optimization. In this formulation, each input-output pair is treated as a goal-reaching task, where $x$ serves as the starting state and $y$ as the intended goal.

Furthermore, we design a binary reward function based on All-Point Satisfaction (APSR) to encourage the action-value network to focus on learning consistent $x-y$ mapping patterns. This design helps prevent structurally different expressions with similar error levels from receiving high rewards simultaneously, which could otherwise misguide the search direction or scatter the optimization trajectory.

In addition, we propose a Structure-Guided Heuristic Exploration (SGHE) strategy, which partitions the search space based on expression structures and allocates independent value networks to distinct structural subspaces, enabling more thorough and efficient exploration across structurally diverse regions.

The main contributions of this work are as follows:
\begin{enumerate}
\item To the best of our knowledge, we propose the first symbolic regression framework based on Goal-Conditioned Reinforcement Learning (GCRL), termed EGRL-SR, which integrates Hindsight Experience Replay to overcome the limitations of error-driven search by introducing a novel paradigm guided by pattern induction.

\item To tackle the ambiguity of error-based reward signals and the imbalance in structural search, we introduce two novel components: the All-Point Satisfaction binary reward function (APSR) and the Structure-Guided Heuristic Exploration strategy (SGHE).

\item Experimental results demonstrate that the proposed method achieves higher recovery rates and greater robustness than existing approaches across multiple public datasets, particularly exhibiting a superior ability to recover more complex expressions under the same search budget. Further ablation studies validate the contributions of the action-value network, the reward function, and the exploration strategy.
\end{enumerate}

\section{Related work}

\subsection{GP-based symbolic regression methods}

In the early development of symbolic regression, Genetic Programming (GP) was one of the dominant approaches, as proposed in Koza \cite{6}. The core procedure involves initializing a population of expression trees, evolving their structures through crossover and mutation, and selecting high-quality individuals for the next generation based on the fitting error.

To improve the search efficiency of traditional GP, various enhancement strategies have been proposed \cite{16,17}. For instance, Virgolin et al. introduced the concept of semantic backpropagation\cite{7}, which guides the evolution of expressions by propagating errors. In a separate work \cite{18}, they proposed a method that evaluates the cooperative relationships among substructures based on error, organizes them into structural groups, and applies localized replacements during crossover to preserve beneficial components and accelerate convergence.

\subsection{EQL-based symbolic regression methods}

Differentiable network-based symbolic regression methods construct multi-layer architectures from predefined operators\cite{9,10}, and jointly optimize both the structure and parameters by minimizing the fitting error via backpropagation. However, these approaches often lack explicit constraints on expression simplicity, making them prone to producing overly complex or redundant models. To mitigate this, researchers have proposed various regularization and pruning strategies\cite{19,20}.

For example, Tsoi et al.\cite{21} proposed a dynamic sparse training method that assigns learnable sparsity thresholds to weights, features, and operators to enable pruning and structural optimization. Li et al. \cite{22}, on the other hand, introduced a reinforcement learning-based approach to dynamically adjust the network architecture, thereby progressively refining the expression generation process.

\subsection{MCTS-based Symbolic Regression methods}

Sun et al.\cite{11}  proposed a symbolic regression framework based on Monte Carlo Tree Search (MCTS), where a reward function balancing fitting accuracy and sparsity is used to guide the search. However, due to the reliance on random strategies during the expansion and simulation phases, the overall search efficiency remains low.

To address this limitation, Kamienny et al. \cite{23} and Li et al. \cite{12} proposed training neural networks to predict the fitness potential of candidate expressions, thereby directing MCTS to expand toward lower-error regions and refining the networks through feedback from the search outcomes. Shojaee et al.\cite{24} further embedded MCTS into the decoding stage of a Transformer model and introduced non-differentiable feedback—such as symbolic complexity penalties—to optimize expression generation, enhancing the model’s capacity for global structural planning.

\subsection{RL-based Symbolic Regression methods}

Symbolic regression tasks are typically limited to input-output pairs without ground-truth expression labels, which renders gradient-based optimization of the generation policy infeasible. To overcome this challenge, reinforcement learning has been increasingly adopted, where fitting error is leveraged as environmental feedback to guide expression generation. A representative approach is Deep Symbolic Regression (DSR) \cite{13}, which employs an RNN-based policy network to construct expressions in a pre-order fashion. At each step, the current partial expression serves as the state, the next symbol selection is treated as the action, and the reward is defined as the inverse of the fitting error.

DSR primarily relies on sampling from the policy network to encourage diversity during exploration, which limits its ability to effectively cover complex expression spaces. To improve global coverage of the symbolic expression space, researchers have proposed hybrid approaches that combine reinforcement learning with genetic programming\cite{14}, or introduced interactive platforms that integrate human feedback\cite{26,27} to compensate for the limitations of purely policy-driven sampling.

\begin{figure*}[t]
\centering
\includegraphics[width=1.7\columnwidth]{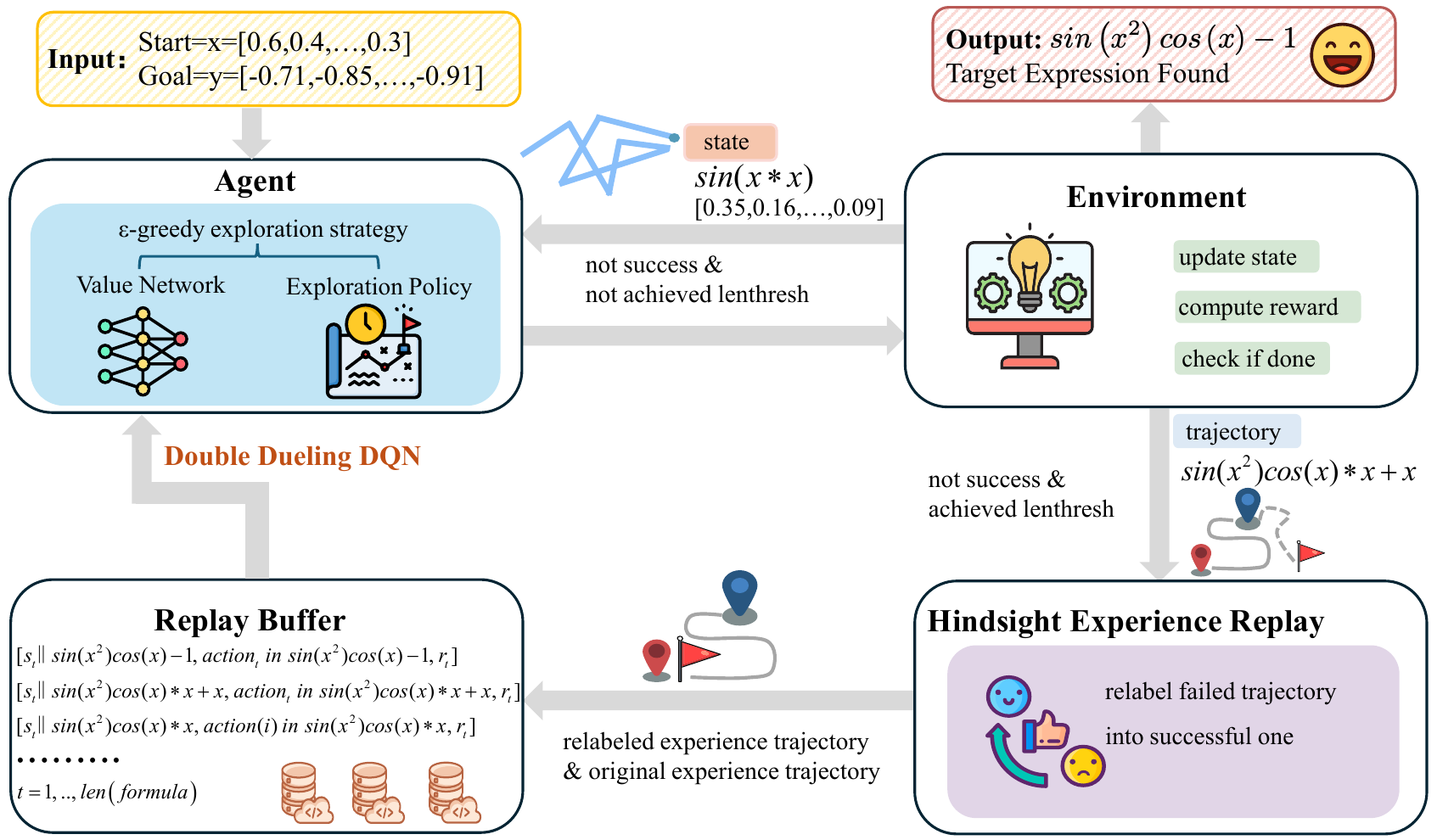} 
\caption{Training framework of EGRL-SR based on agent–environment interaction}
\label{figure1}
\end{figure*}

\begin{figure*}[t]
\centering
\parbox{0.36\linewidth}{
  \centering
  \includegraphics[width=\linewidth]{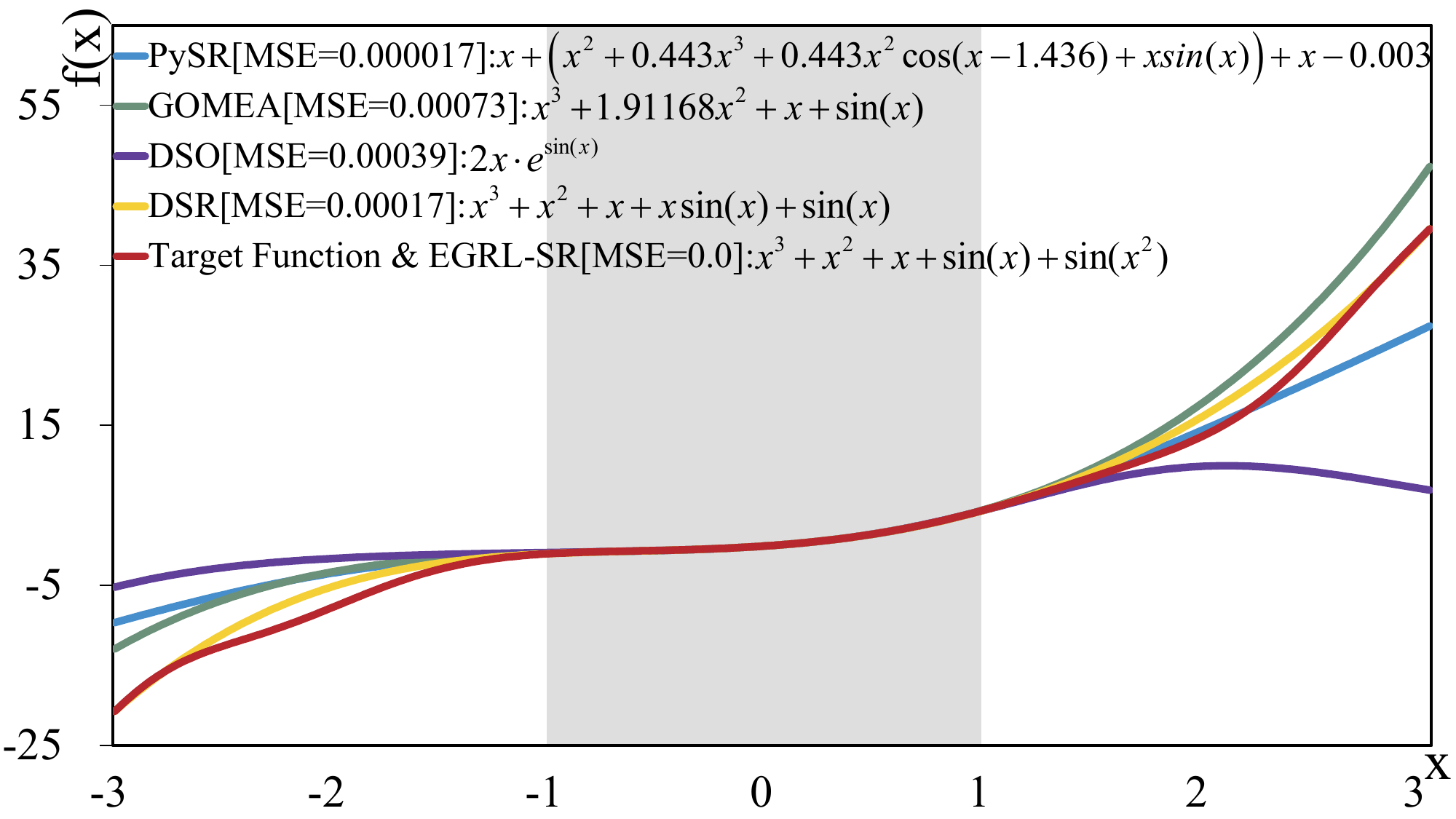}
  \caption{Function plots of the target expression and the corresponding expressions generated by SR algorithms (EGRL-SR, DSR, DSO, GOMEA, PySR). Twenty training points are uniformly sampled from the interval $[-1, 1]$, and the mean squared error (MSE) quantifies the fitting accuracy within this interval. To assess the generalization capability beyond the original input range, the plotting range is extended to $[-3, 3]$.}
  \label{figure2}
}
\hfill
\parbox{0.3\linewidth}{
  \centering
  \includegraphics[width=\linewidth]{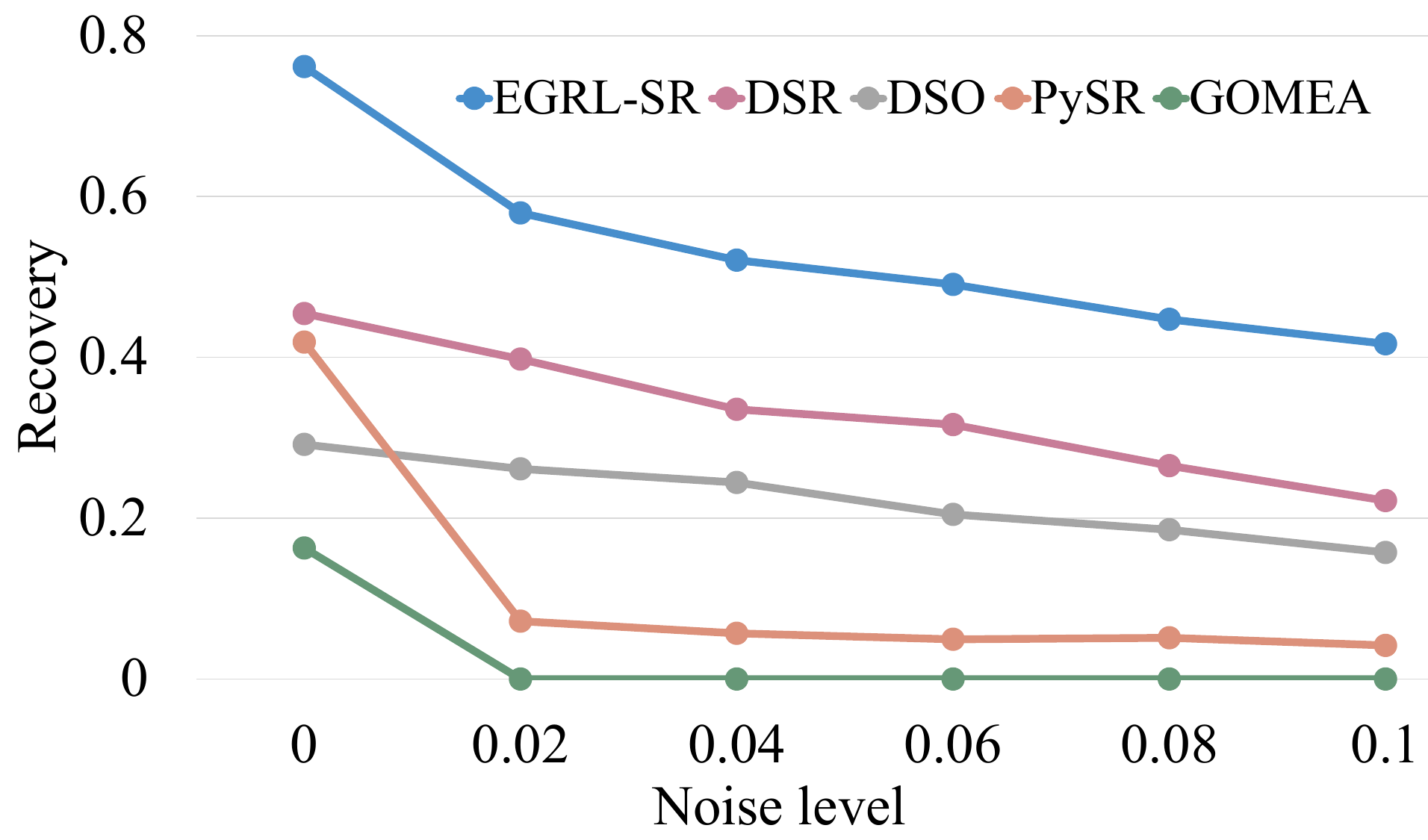}
  \caption{Expression recovery rates of EGRL-SR and baseline methods under different noise levels. The recovery rates of Pretraining-based methods (NeSymReS, E2E), Monte Carlo tree search method (TPSR), and Equation Learner methods (EQL-Div, DySymNet) remain at 0\%.}
  \label{figure3}
}
\hfill
\parbox{0.3\linewidth}{
  \centering
  \includegraphics[width=\linewidth]{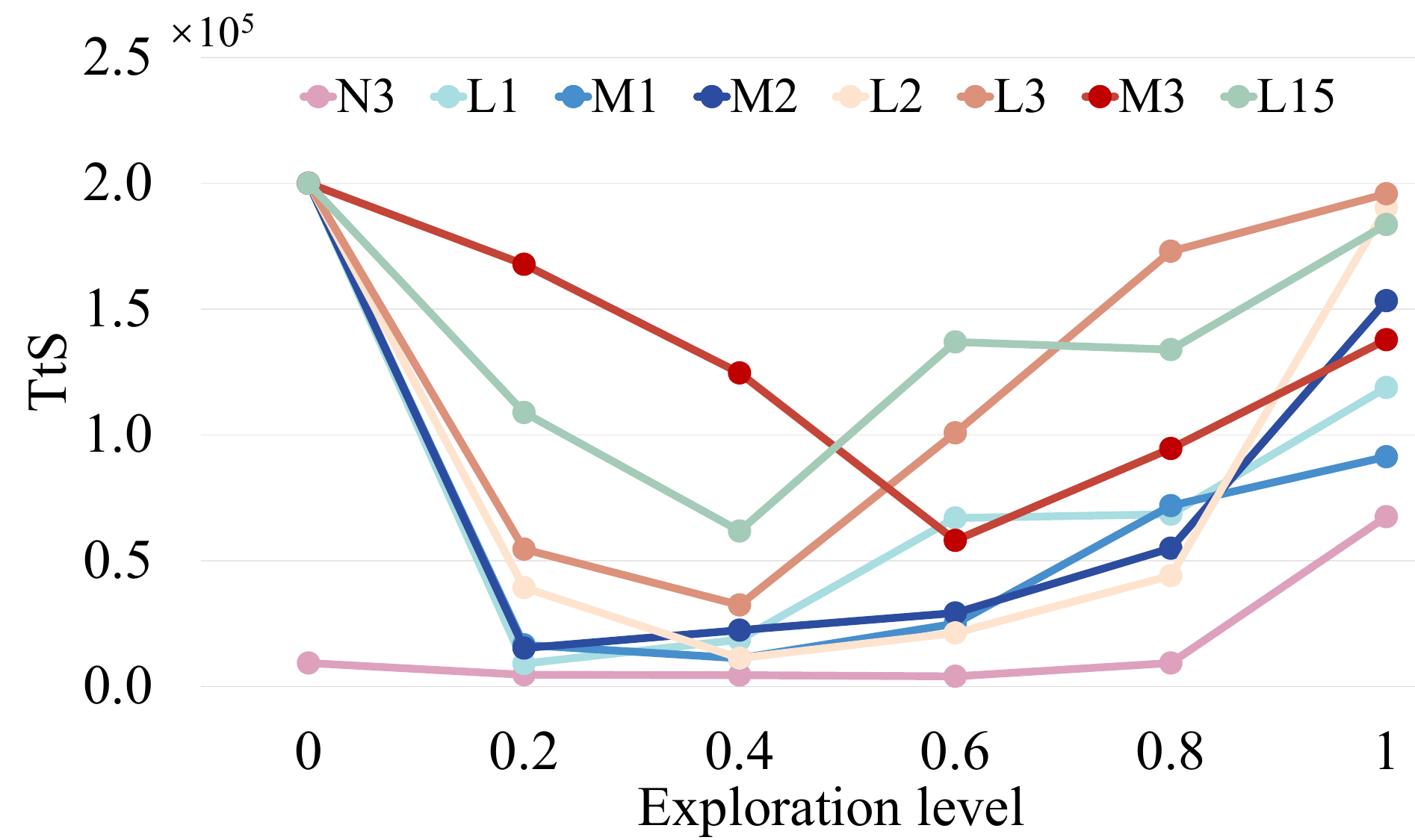}
  \caption{Average TtS for similar-length expressions with different structural complexities under varying $\epsilon$ values. Expressions starting with “N” and “L” come from the Nguyen and Livermore datasets, respectively, with the number indicating their index. “M”-prefixed expressions are manually designed to increase structural diversity at the similar length.}
  \label{figure4}
}
\end{figure*}

\section{Methods}

This chapter is organized as follows: we begin by formalizing the symbolic regression task as a Goal-Conditioned Markov Decision Process (GC-MDP), explicitly defining the state space, action space, and reward function. We then describe the key components of our framework, including the Hindsight Experience Replay (HER) , the optimization algorithm for the action-value network (Double Dueling DQN), and the Structure-Guided Heuristic Exploration (SGHE) strategy.

\subsection{Goal-Conditioned Markov Decision Process}

As illustrated in Figure \ref{figure1}, symbolic regression training is framed as an agent–environment interaction process. At each step, the agent follows an $\epsilon$-greedy strategy for action selection: with probability $1 - \epsilon$, it exploits prior experience by choosing actions based on the value network; with probability $\epsilon$, it explores the expression space using the Structure-Guided Heuristic Exploration (SGHE) strategy. The environment then returns a new state and evaluates whether the target has been reached: if so, the trajectory is marked as successful; if not and the expression length remains within the threshold, the agent proceeds to generate further tokens; if the threshold is reached, the failed trajectory is reinterpreted as successful via Hindsight Experience Replay (HER). The agent subsequently samples trajectories from the replay buffer to train the value network using a reinforcement learning algorithm. The complete workflow of the algorithm is presented in Appendix A.

\subsubsection{State}

The state $s$ captures the agent’s numerical progress during the expression construction process. It is defined as the concatenation of the current intermediate numeric output (denoted as $x_{\text{now}}$) and the target output $y$, i.e., $s = x_{\text{now}} \| y$. For single-variable tasks, the initial state is $s_0 = x \| y$; for multi-variable tasks, $s_0 = 0 \| y$. When the state reaches $y \| y$, it indicates that the agent has successfully constructed an expression whose numeric evaluation perfectly matches the target $y$. The sequence of actions along this trajectory constitutes the postfix representation of the corresponding expression.

\subsubsection{Action}

The action space comprises variables, unary operators, and binary operators. At each time step, the agent can choose a variable operate to generate a new node, apply a unary operator to transform an existing node, or use a binary operator to combine two isolated nodes. We adopt a postfix generation strategy, where variables and operands are placed before operators. This design enables the agent to observe its intermediate numeric output at each step and adapt the generation path dynamically based on its proximity to the target.

\subsubsection{Reward}

In this study, we propose a sparse binary reward mechanism called the All-Point Satisfaction Reward (APSR). Under this scheme, the agent receives a reward of 1 only if the constructed expression meets a predefined accuracy threshold across all input samples; otherwise, the reward is 0.

The APSR is designed to overcome two fundamental limitations of conventional continuous reward functions based on fitting error. First, a low fitting error does not necessarily imply structural correctness—expressions that are structurally flawed but highly expressive may still achieve low errors, leading the policy to converge to suboptimal solutions. Second, structurally different expressions may yield similar error values, making it difficult for error-based rewards to distinguish them effectively. This lack of structural guidance hampers policy optimization and poses challenges to stable convergence.

\subsection{Hindsight Experience Replay}
In actual search processes, trajectories are guided solely by the current target $y$, and the replay buffer lacks successful experiences associated with alternative targets $y’$. This hinders the action-value network from learning generalizable construction patterns across multiple goals, thus limiting its generalization capacity. To address this issue, we introduce Hindsight Experience Replay (HER), which replaces the original target $y$ with an intermediate output actually reached by the agent, thereby yielding successful training samples and providing positive reward signals.

Specifically, HER selects several intermediate outputs from each failed trajectory as new goals, recalculates the associated rewards, and adds the relabeled samples to the experience buffer for subsequent learning by the action-value network. This allows the network to encounter a diverse set of goal-conditioned construction paths during training, extract generalizable $x-y$ mapping patterns, and ultimately improve its performance in reconstructing the current target. A detailed description of the algorithm can be found in Appendix A.

\subsection{Reinforcement Learning Algorithm — Double Dueling DQN}

Since HER violates the assumption that training data must be generated by the current policy, it is incompatible with on-policy algorithms\cite{30,31}. To address this, we adopt the off-policy Double Dueling DQN\cite{32,33,34}, which can effectively leverage both historical trajectories and relabeled experiences.

Double Dueling DQN enhances Q-value estimation by introducing a decomposition into a state-value function $V(s)$ and an advantage function $A(s, a)$, which allows the model to more accurately differentiate between state value and action-specific advantage:
\begin{equation}
Q(s,a) = V(s) + \left(A(s,a) - \frac{1}{|\mathcal{A}|} \sum_{a’} A(s,a’)\right)
\end{equation}
In addition, the algorithm employs a double-network architecture that separates action selection from value evaluation, effectively mitigating the overestimation bias common in traditional Q-learning. The target value is computed as:
\begin{equation}
y = r + \gamma Q_{\text{target}}(s’, \arg\max_{a’} Q_(s’, a’))
\end{equation}
The network is trained by minimizing the following loss:
\begin{equation}
\mathcal{L}(\theta) = {E}_{s, a} \left[ \left( y - Q(s, a; \theta) \right)^2 \right]
\end{equation}

The target network parameters are updated via soft updates, gradually tracking the online network with a small step size.

	\begin{table*}[htbp]
		\footnotesize
		\centering
		\renewcommand{\arraystretch}{1.4}
		\setlength{\tabcolsep}{0.5mm}
		
		\begin{tabular}{
				|>{\centering\arraybackslash}m{1.6cm}
				|>{\centering\arraybackslash}m{1.6cm}
				|>{\centering\arraybackslash}m{1.6cm}
				|>{\centering\arraybackslash}m{1.6cm}
				|>{\centering\arraybackslash}m{1.6cm}
				|>{\centering\arraybackslash}m{1.6cm}
				|>{\centering\arraybackslash}m{1.3cm}
				|>{\centering\arraybackslash}m{1.3cm}
				|>{\centering\arraybackslash}m{1.3cm}
				|>{\centering\arraybackslash}m{1.3cm}
				|>{\centering\arraybackslash}m{1.3cm}
				|}
			\hline
			{\centering \textbf{Expression}\\\textbf{length}\\\textbf{(number)}} &
			\textbf{EGRL-SR} &
			\textbf{DSR} &
			\textbf{PySR} &
			\textbf{GOMEA} &
			\textbf{DSO} &
			\textbf{TPSR} &
			{\centering \textbf{NeSym}\\\textbf{ReS}} &
			{\centering \textbf{End to}\\\textbf{End}} &
			\textbf{EQL-Div} &
			{\centering \textbf{DySym}\\\textbf{Net}} \\
			\hline
			$\leq 8$(4)       & 100\% (0\%)& 96\%(9\%)& 100\%(0\%) & 40\%(49\%) & 98\%(4\%)& 0\%(0\%)& 0\%(0\%)& 0\%(0\%)& 0\%(0\%) & 0\%(0\%) \\
			\hline
			9--10(4)        & 100\%(0\%) & 100\%(0\%) & 100\%(0\%) & 50\%(58\%) & 77\%(28\%) & 0\%(0\%) & 17\%(19\%) & 0\%(0\%) & 0\%(0\%)  & 0\%(0\%) \\
			\hline
			11--12(4)       & 100\%(0\%) & 77\%(32\%)  & 46\%(37\%)  & 11\%(21\%) & 25\%(18\%) & 0\%(0\%) & 0\% (0\%)& 0\%(0\%)  & 0\%(0\%)  & 0\%(0\%) \\
			\hline
			13--14(6)       & 100\%(0\%) & 61\%(49\%)  & 47\%(39\%)  & 26\%(33\%) & 29\%(38\%) & 0\%(0\%) & 0\%(0\%) & 0\%(0\%)  & 0\%(0\%)  & 0\%(0\%) \\
			\hline
			15--16(10)       & 88\%(17\%)  & 29\%(42\%)  & 30\%(36\%)  & 8\%(21\%)  & 16\%(33\%) & 0\%(0\%) & 0\%(0\%) & 0\%(0\%)  & 0\%(0\%) & 0\%(0\%)\\
			\hline
			17--20(4)       & 56\%(38\%)  & 11\%(13\%)  & 48\%(41\%) & 11\%(21\%) & 2\%(4\%) & 0\%(0\%) & 0\%(0\%)& 0\%(0\%) & 0\%(0\%) & 0\%(0\%)\\
			\hline
			21--30(5)       & 58\%(50\%)  & 23\%(43\%)  & 0\%(0\%)   & 0\%(0\%)  & 17\%(37\%) & 0\%(0\%) & 0\%(0\%) & 0\%(0\%)  & 0\%(0\%)  & 0\%(0\%) \\
			\hline
			$\geq31$(5)     & 0\%(0\%)   & 0\%(0\%)   & 0\%(0\%)   & 0\%(0\%)  & 0\%(0\%)  & 0\%(0\%) & 0\%(0\%) & 0\%(0\%)  & 0\%(0\%)  & 0\%(0\%) \\
			\hline
		\end{tabular}
		\caption{Average exact recovery rates of EGRL-SR and baseline methods on all expressions within each expression length interval. The first column indicates the expression length interval (with the number of expressions in each interval in parentheses). The remaining columns present the average exact recovery rates (with standard deviation in parentheses) for each algorithm on all expressions within the corresponding interval.
}
\label{table1}
	\end{table*}

\subsection{Structure-Guided Heuristic Exploration Strategy}

Given that EGRL-SR employs an off-policy reinforcement learning algorithm capable of flexibly incorporating external trajectories, we propose a Structure-Guided Heuristic Exploration (SGHE) strategy. SGHE partitions the expression space based on multiple structural attributes, establishing a set of independent exploration directions. Each direction is equipped with its own action-value network and dedicated replay buffer to enable targeted and structurally aware learning.

This strategy promotes balanced exploration across structurally diverse expressions, mitigating the bias toward low-error expressions confined to specific structural patterns. It facilitates the discovery of target expressions that are structurally simple yet exhibit hard-to-model $x-y$ mappings.

In contrast to on-policy reinforcement learning-based symbolic regression methods, EGRL-SR supports external structural intervention and directional control over the exploration process, enabling a tight coupling between symbolic expression exploration and learning. A detailed theoretical analysis is provided in Appendix B.

Specifically, each exploration direction is defined by a combination of three structural attributes: the number of unary operators, the nesting depth, and the length range of the sub-expression to which the unary operator is applied. 

With probability $\epsilon$, the agent adheres to the predefined structural constraints—proceeding to generate variable and binary operator nodes, and applying the unary operator only when the sub-expression reaches the required length.

With probability $1 - \epsilon$, the agent selects the next action purely based on the value network, unconstrained by any external structural heuristics, thereby making full use of accumulated experience.

\section{Experiments}
This paper investigates the following five research questions through comparative and ablation studies:

RQ1: Given the same search budget, can EGRL-SR reliably recover expressions that are typically difficult for existing symbolic regression methods to discover?

RQ2: Can EGRL-SR accurately recover expressions under noisy conditions by consistently identifying the underlying $x-y$ mapping patterns?

RQ3: Can the value network trained from experience effectively guide the expression search process?

RQ4: Can SGHE improve coverage of the structural search space and facilitate the discovery of structurally simple expressions with hard-to-learn $x-y$ mappings?

RQ5: Can APSR prevent the value network from prematurely converging to suboptimal expressions due to structurally incorrect formulas with similarly low error?

\subsection{Dataset}

To systematically evaluate the recovery performance of the proposed method across varying structural patterns and levels of complexity, we adopt three widely used symbolic regression benchmarks: Nguyen\cite{35}, Livermore\cite{14}, and Keijzer\cite{36}. To stratify the recovery difficulty, all target expressions from these datasets are consolidated and ranked by increasing expression length. Details regarding the expression forms, lengths, sources, and corresponding input sampling ranges contained in the dataset can be found in Appendix C. 

\subsubsection{Target Expressions for Ablation study}

To assess the critical contributions of the action-value network, exploration strategy, and reward mechanism in guiding expression construction, we conduct ablation experiments on eight target expressions that share similar lengths but differ significantly in structural complexity (as shown in table \ref{table2}). Among them are three synthetically constructed expressions designed to increase structural complexity. By standardizing expression length, we control for variations in search space size—ensuring that any failure in recovery can be attributed to limitations in structural pattern recognition rather than the confounding effect of a large search space.

\subsection{Comparison Algorithms}

To comprehensively assess the performance of our proposed method, we compare it against a diverse set of state-of-the-art symbolic regression baselines, spanning the major methodological paradigms: Reinforcement Learning (DSR)\cite{13}, Pretraining-based approaches (NeSymReS, E2E)\cite{4,3}, Genetic Programming (GOMEA, PySR)\cite{18,17}, Equation Learner method (EQL-Div, DySymNet)\cite{10,22}, Monte Carlo Tree Search (TPSR)\cite{24}, and RL-GP hybrid method (DSO)\cite{14}. Details of each baseline are provided in Appendix D.

\subsection{Evaluation Metric}

While many studies adopt $R^2$ as the primary evaluation metric, this often incentivizes the generation of overly complex expressions that boost accuracy through structural complexity rather than genuine model effectiveness. To address this issue, we instead adopt the exact recovery rate of the target expression—including constants—as the evaluation metric.
A unified search budget of 1.6 million steps is allocated to each algorithm (corresponding to 8 search directions with 200,000 steps each), and each target expression is evaluated over 12 independent trials. All algorithms adopt a unified operator set $\left\{ +,-,\ast ,/,sin,cos,exp,log \right\}$ for expression recovery.

\subsection{Expression Recovery Rate Evaluation - RQ1}
To evaluate the recovery performance and comparative advantage of the proposed method across expressions of varying complexity, we conduct a series of systematic comparative experiments. Details on hyperparameter settings and the computational environment are provided in Appendix E.

As shown in table \ref{table1}, the proposed method exhibits strong recovery performance across most expression length ranges, with its overall average recovery rate consistently exceeding those of the baseline methods.

Figure \ref{figure2} illustrates the expressions generated by the five best-performing algorithms, along with their corresponding function plots, when the target expression is $x_1^3 + x_1^2 + x_1 + \sin(x_1^2) + \sin(x_1)$. It can be observed that only EGRL-SR successfully recovers the exact expression, while the expressions identified by the other methods—despite achieving very low fitting errors—exhibit structural deviations from the ground-truth target.

The results indicate that EGRL-SR, by leveraging structural guidance and experience-driven mechanisms, effectively mitigates the bias caused by error-dominated search and enhances the model’s ability to identify underlying $x-y$ mappings, thereby improving expression recovery performance. In contrast, other search-based symbolic regression methods tend to overly rely on fitting error, making them susceptible to structurally flawed expressions with deceptively low fitting errors that represent local optima.

The exact recovery rates of symbolic regression methods based on Pretraining, Monte Carlo Tree Search, and Equation Learner method remain close to zero across different levels of expression complexity. This is likely because pretraining-based approaches rely on synthetic datasets to train expression-generating networks; however, due to the virtually unbounded nature of the expression space, the training data cannot adequately capture structural diversity, limiting the model’s ability to learn generalizable symbolic construction rules. MCTS-based methods, which rely on pretrained networks during node expansion, suffer from similar distributional mismatch. The suboptimal performance of the Equation Learner method may stem from the absence of effective structural pruning strategies and regularization mechanisms. Per-expression recovery results for all algorithms are detailed in Appendix F.

\subsection{Robustness to Noise - RQ2}

In real-world applications, observational data often contain noise, and thus the goal of symbolic regression is to recover an underlying generalizable function from such perturbed data. To evaluate whether EGRL-SR remains capable of identifying the core $x-y$ mapping patterns under noisy target outputs, we conducted a systematic noise experiment. 

Specifically, we added independent Gaussian noise to the dependent variable in the training data, with zero mean and a standard deviation proportional to the root mean square of $y$, gradually increasing the noise level from $0$ to $10^{-1}$. The datasets and search budget are consistent with those used in the recovery rate experiments.

As illustrated in Figure \ref{figure3}, EGRL-SR consistently achieves superior recovery rates across all noise levels in the expression recovery experiments. Reinforcement learning-based methods such as DSR and DSO maintain relatively stable performance, with recovery rates declining gradually as noise increases—indicating a degree of robustness. However, their overall performance remains inferior to that of EGRL-SR. In contrast, genetic programming methods like PySR and GOMEA suffer a sharp drop in performance upon the introduction of noise, with recovery rates approaching 0\%. This is likely because they rely heavily on fitting error as the optimization signal, making them prone to overfitting noisy data.

\begin{table}[htbp]
	\footnotesize
	\centering
	\renewcommand{\arraystretch}{1.4}
	\setlength{\tabcolsep}{0.2mm}
	
	\begin{tabular}{
			|>{\centering\arraybackslash}m{4.0cm}
			|>{\centering\arraybackslash}m{0.9cm}
			|>{\centering\arraybackslash}m{0.9cm}
			|>{\centering\arraybackslash}m{0.9cm}
			|>{\centering\arraybackslash}m{0.9cm}|}
		\hline
		{\centering \textbf{Target Expression}} &
		{\centering \textbf{HER}\\\textbf{APSR}\\\textbf{SGHE}} &
		{\centering \textbf{HER}\\\textbf{APSR}\\\textbf{RE}} &
		{\centering \textbf{HER}\\\textbf{nRR}\\\textbf{SGHE}} &
		{\centering \textbf{nRR}\\\textbf{SGHE}} \\
		\hline
		$N3: x_{1}^5 + x_{1}^4 + x_{1}^3 + x_{1}^2 + x_{1}$&100\%& 100\%&100\%&100\% \\
		\hline
		$L1: \sin(x_{1}^2) + x_{1} + \frac{1}{3}$ &100\%&75\%&83\%&8\%     \\
		\hline
		$M1: \sin[(x_{1} - 1)^2] + x_{1}^2$  & 100\%& 8\%&100\%&100\% \\
		\hline
		$M2: \log(x_{1}^5 + 2x_{1}^3 + x_{1})$  &100\%&0\%&83\%& 58\%     \\
		\hline
		$L2: \sin(x_{1}^2) \cos x_{1} - 2$   & 100\% & 67\% &100\%&100\% \\
		\hline
		$L3: \sin(x_{1}^3)  \cos(x_{1}^3) - 1$    & 100\%&25\%&67\%&42\% \\
		\hline
		$M3: \sin(x_{1}^2 + 1) \cos(x_{1}^2) + x_{1}$  &100\%&0\%&33\%&33\% \\
		\hline
		$L15: x_{1}^{1/5}=exp\left( \frac{1}{5} log\left( x_{1} \right) \right)$   & 83\%& 0\%&50\%&8\%  \\
		\hline
	\end{tabular}
	\caption{Exact recovery rates for similar-length expressions with different structural complexities under four configurations: (a) the full method (HER + APSR + SGHE); (b) replacing the structure-guided heuristic exploration (SGHE) with random exploration (HER + APSR + RE); (c) replacing APSR with a continuous reward based on normalized RMSE (HER + nRR + SGHE); (d) further removing HER from configuration c (nRR + SGHE). The expression indices (L1, L2, etc.) used in this table correspond to those defined in Figure \ref{figure4}.
}
\label{table2}
\end{table}

\subsection{Ablation experiment}

\subsubsection{Effectiveness of Action-Value Network in Guiding Search - RQ3}

To evaluate the search guidance capability of the action-value network, we conducted a controlled experiment by introducing perturbations to the $\epsilon$-greedy exploration strategy. If the network provides effective guidance, a lower $\epsilon$ value should result in a reduced average number of Trajectories-to-Success (TtS). Specifically, TtS refers to the average number of searches—within the exploration direction where the target expression is successfully recovered—required by the action-value network to recover the expression. A smaller TtS indicates higher search efficiency.

As shown in Figure \ref{figure4}, the TtS decreases markedly as $\epsilon$ decreases from 1, with the lowest value observed around $\epsilon$=0.2 or 0.4. This trend indicates that the increasing guidance provided by the value network leads to improved search efficiency, thereby verifying its essential role in directing the search process.

In addition, Figure \ref{figure4} shows that polynomial expressions (pink line) are the most readily recovered, whereas target expressions involving two unary operators applied to long subexpressions (green and dark orange lines) are the most difficult to recover. This suggests that even when the solution space size is fixed, different expression structures pose significantly different challenges for search and learning. Given the high structural diversity of expressions, nearly every structure type may contain expressions with low fitting error. If the search is guided solely by fitting error, the algorithm tends to favor expressions with lower error in its preferred structures, thereby overlooking the structurally correct ground-truth expressions.

\subsubsection{Evaluation of Structural Exploration Coverage - RQ4}
To evaluate whether SGHE can improve structural space coverage to discover target expressions that are structurally simple yet challenging to recover due to complex $x-y$ mapping patterns, we conduct a comparative analysis of the recovery performance of SGHE versus random exploration across structurally diverse expressions.

As illustrated in table \ref{table2}, given equal expression lengths, the random exploration strategy can only recover expressions with specific structural patterns, suggesting that the action-value network is biased toward certain structures and tends to be trapped in local optima in the absence of structural guidance.

\subsubsection{Evaluation of Reward Function Robustness Against Misguidance - RQ5}

To evaluate whether the proposed All-Point Satisfaction Reward (APSR) can guide the policy toward learning structurally diverse patterns and mitigate the tendency of the model to converge on structurally incorrect yet low-error expressions, we conducted two sets of comparative experiments.

The first experiment replaces APSR with a continuous reward function based on the normalized root mean square error (nRMSE), formulated as $r = \frac{1}{1 + \text{nRMSE}}$.
The second experiment further removes HER on top of this setup to more closely align with the conventional error-driven search paradigm.

As shown in table \ref{table2}, replacing APSR with reward based on nRMSE leads to a marked decline in overall recovery performance. This decline is likely due to structurally incorrect expressions with low fitting errors receiving reward signals comparable to those of correct trajectories generated by HER, thereby misleading the optimization of the value network.

Upon the further removal of HER, performance deteriorates even more significantly. This is likely because the absence of goal reconstruction deprives the agent of informative trajectories, making it difficult for the value network to capture the mapping patterns between $x$ and diverse target outputs $y’$. As a result, training is more easily biased by misleading error feedback.

A few expressions remain unaffected, possibly because the structural subspace defined by SGHE inherently contains few low-error expressions within the given value range.

\section{Conclusion and Future Work}

This study introduces a novel symbolic regression framework grounded in goal-conditioned reinforcement learning, referred to as EGRL-SR, which establishes a new search paradigm guided by structural pattern induction. However, the current approach represents constants indirectly via combinations of variables and operators, which substantially increases the search space when handling expressions with complex constant terms, thereby impairing algorithm performance. Future work will focus on developing effective strategies for recovering expressions involving complex constants within the GCRL framework.

\bibliography{aaai2026_my}


\end{document}